\documentclass[runningheads]{llncs}
\usepackage{graphicx}
\usepackage{booktabs} 
\usepackage{multirow}
\usepackage{threeparttable}
\usepackage{verbatim}
\usepackage{multicol}
\usepackage{algorithm}
\usepackage{algorithmic}
\usepackage{underscore}
\usepackage{amsmath}
\usepackage{bm}
\usepackage{afterpage}
\usepackage{wrapfig}
\usepackage{subfigure}
\usepackage{lipsum}

\usepackage{amssymb}
\usepackage{marvosym}

\begin{document}
\title{Small Temperature is All You Need for Differentiable Architecture Search}
\titlerunning{Small Temperature is All You Need for DARTS}
\author{Jiuling Zhang\inst{1,2}\and
Zhiming Ding\inst{2,1}\textsuperscript{\Letter}}
\institute{University of Chinese Academy of Sciences, Beijing 100049, China \and
Institute of Software, Chinese Academy of Sciences, Beijing 100190, China\\
\email{zhangjiuling19@mails.ucas.ac.cn}, \email{zhiming@iscas.ac.cn}}
%
%
%
%
%
\maketitle              
\begin{abstract}
Differentiable architecture search (DARTS) yields highly efficient gradient-based neural architecture search (NAS) by relaxing the discrete operation selection to optimize continuous architecture parameters that maps NAS from the discrete optimization to a continuous problem. DARTS then remaps the relaxed supernet back to the discrete space by one-off post-search pruning to obtain the final architecture (finalnet). Some emerging works argue that this remap is inherently prone to mismatch the network between training and evaluation which leads to performance discrepancy and even model collapse in extreme cases. We propose 
to close the gap between the relaxed supernet in training and the pruned finalnet in evaluation through utilizing small temperature to sparsify the continuous distribution in the training phase. To this end, we first formulate sparse-noisy softmax to get around gradient saturation. We then propose an exponential temperature schedule to better control the outbound distribution and elaborate an entropy-based adaptive scheme to finally achieve the enhancement. We conduct extensive experiments to verify the efficiency and efficacy of our method.

\end{abstract}
\section{Introduction}
DARTS abstracts the search space as a cell-based directed acyclic graph composed by $V$ nodes $H = \{ {h_1},{h_2},...,{h_V}\} $ and compound edges $C = \{ c_{{1,2}},...,c_{{V - 1,V}}\} $. Every node represents feature maps and each edge subsumes all operation candidates to express the transformations between nodes. Compound edge ${c_{u,v}}$ connects node $u$ to $v$ and associates three attributes: candidate operation set ${O_{c}} = \left\{ {o_{c}^1,o_{c}^2,...,o_{c}^M} \right\}$, corresponding operation parameter set ${A_{c}} = \left\{ {a _{c}^1,a _{c}^2,...,a _{c}^M} \right\}$, probability distribution of the parameters ${\beta_{c}} = {\rm{softmax}}({A_{c}})$.
Every intermediate node is densely-connected to all its predecessor through an edge ${h_v} = {c_{u,v}}({h_u})$ and weighted product sum ${c_{u,v}}({h_u}) =  \left\langle {\beta_{c}},{O_{c}}({h_u}) \right\rangle $ for $u<v$.  
Generally, a unified set of operation candidates $O = \left\{ {o^1,o^2,...,o^M} \right\}$ is defined for all edges in the space. 
The network that encodes all architectural candidates is termed supernet. 
In the training phase, DARTS first divides data into training and validation sets and then formulates a bilevel objective depicted in Eq.(1) to alternately optimize architecture parameter $a$ and operation weight $\omega$ on the validation set and training set respectively. We refer to \cite{liu2018darts} for more details of DARTS.
\begin{equation}
\mathop {\min }\limits_a  {L_{val}}({\omega ^ * }(a ),a ){\rm{\ \ s}}{\rm{.t}}{\rm{.\ }}{\omega ^ * } = \arg \mathop {\min }\limits_\omega  {L_{train}}(\omega ,a )
\end{equation}
Henceforth, we abbreviate operation weight $\omega$ as weights and architecture parameter $a$$\in$$A$ as parameters. After the training phase, DARTS selects the operation associated with the largest entry (probability component after normalized by softmax) through a post-search pruning depicted in Eq.(2) to obtain the final architecture (finalnet) for evaluation. In sum, the supernet is trained in a multi-path manner while the finalnet is evaluated in a single-path manner \cite{luo2022surgenas}.
\begin{equation}
{o^{i}_c} \in O\ {\rm{for}} \ c\in C\ {\rm{where}}\ i={{\arg\max} _{i\in\{1,...,M\}}}\ \beta_c,\ {\beta}_{c} = {\rm{softmax}} ({A_{c}})
\end{equation}
where the operation ${o}^{i}_c$ is selected on edge $c$ due to the largest entry $\beta^i_c$ in $\beta_c$. 
However, the parameter-value-based post-search pruning depicted in Eq.(2) inevitably risks mismatching the architecture between supernet and finalnet which ultimately leads to performance discrepancy between training and evaluation in DARTS. Gradient-based NAS methods are deemed to favor architectures that are easier to be trained~\cite{shu2019understanding}. More skip connections can obviously help the network to converge faster which is considered as an unfair advantage in an exclusive competition~\cite{chu2020fair,chu2020darts}. This unfairness leads to an abnormal preference for the skip connection in some cases during optimization. The architecture mismatch exacerbates this issue since the one-off post-pruning generally discards all operations except the dominant one ($O\setminus o^{\arg{\max }_i (\beta)}$). In extreme cases, the pruning removes all operations except the skip connection which causes collapse or the catastrophic failure of DARTS \cite{arber2020understanding,dong2019bench,liang2019darts+,cheng2020hierarchical}. 

Both Fair-DARTS \cite{chu2020fair} and GAEA \cite{li2021geometry} pointed out that the sparse parameters are crucial to alleviating the architecture mismatch. In particular, GAEA emphasized that obtaining sparse final architecture parameters is critical for good performance, both for the mixture relaxation, where it alleviates the effect of overfitting, and for the stochastic relaxation, where it reduces noise when sampling architectures. In DARTS, \textit{the sparse parameters refer to the sparse distribution $\beta$ (low entropy) after the parameters normalized by softmax}, i.e. $\beta = {\rm{softmax}} (A)$. Sparse parameters intrinsically reduce the gap between multi-path supernet and single-path finalnet as shown in Figure~\ref{fig1}.
\cite{wu2021neural} proposed to combine single-path and multi-path space by a Sparse Group
Lasso constraint but impose non-trivial additional time consumption due to it's harder to converge. Temperature coefficient is widely used to control the sparseness and smoothness of the (gumbel) softmax output. 
In this paper, we propose to achieve sparse training straightforwardly by employing a small temperature on softmax. Our contributions can be summarized:
\begin{itemize}
\setlength{\itemsep}{0pt}
\setlength{\parsep}{0pt}
\setlength{\parskip}{0pt}
\item We propose sparse-noisy softmax (sn-softmax) to alleviate the gradient saturation which causes premature convergence of the training of parameters while utilizing small temperature in the training phase;
\item We propose exponential temperature schedule (ETS) and entropy-based adaptive scheme to maintain and better control the sparsity of $\beta$ that inherently narrows the gap between the relaxed supernet in training and the pruned finalnet in evaluation;
\item We carry out extensive evaluations on multiple spaces and datasets and conduct further ablations to show the effect of different hyperparameter choices.
\end{itemize}

\section{Methodology}
Utilizing small temperature to sparsify $\beta$ in training is non-trivial for DARTS because the gradient saturation will impede the propagation when softmax converges. Likewise, the operation weights will not be updated either when the operation output are weighted by a zero entry in $\beta$. If that happens in the middle training of DARTS, supernet converges prematurely. In this section, we start with formulating sparse-noisy softmax to alleviate the gradient saturation when the outbound $\beta$ converges. After that, we propose an exponential temperature schedule (ETS) to better control the temperature $t$ to smooth the swing of the outbound $\beta$ in training. We provide an \textbf{e}ntropy-based \textbf{d}ynamic \textbf{d}ecay (EDD) to finally realize a flexible and robust enhancement for DARTS.

\subsection{Sparse-noisy Softmax}
\begin{wrapfigure}[]{r}{0.55\textwidth}
\begin{center}
\vskip -0.8in
\includegraphics[width=0.55\textwidth]{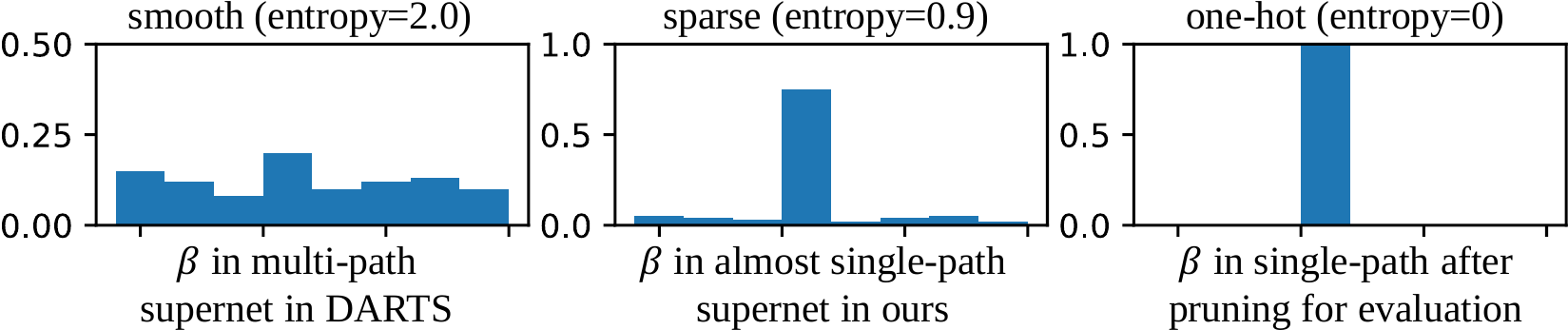}
\vskip -0.1in
\caption{Multi-path, sparse and single-path.}
\label{fig1}
\end{center}
\vskip -0.55in
\end{wrapfigure}

Softmax normalizes the input vector ${A} = \{ {a^1},...,{a^M}\}$ to a probability distribution ${\beta} = \{ {\beta^1},...,{\beta^M}\}$ depicted in Eq.(3).
\begin{equation}
{\rm softmax} (\frac{A}{t}) = {\beta_t}\ \ {\rm{where}}\ \beta_t^i=\frac{\exp(a^i/t)}{\sum_{j=1}^{M}{\exp(a^j/t)}}
\end{equation}
where $t$ is the temperature coefficient and $M$ is the total entries of the input of softmax in DARTS. The smaller the temperature $t$, the sharper the outbound $\beta$ is and the closer the distribution compared to post-search one-hot argmax pruning in Eq.(2). The derivative of softmax can be gotten through:
\begin{equation}
\frac{{\partial \beta_t^i}}{{\partial {a^i}}} = \left\{ {\begin{array}{*{20}{c}}
{\frac{{\beta_t^i(1 - \beta_t^i)}}{t}\ \ {\rm{for}}\ i = j}\\
{\frac{{ - \beta_t^i\beta_t^j}}{t}\ \ {\rm{for}}\ i \ne j}
\end{array}} \right.
\end{equation}
by which we can get the Jacobian matrix as:
\begin{equation}
\frac{{\partial {{\beta}_t}}}{{\partial {A}}} = \frac{1}{t}\left[ {\begin{array}{*{20}{c}}
{\beta_t^1 - {{[\beta_t^1]}^2}}&{ - 
\beta_t^1\beta_t^2}&{ - \beta_t^1\beta_t^3}& \cdots &{ - \beta_t^1\beta_t^M}\\
{ - \beta_t^2\beta_t^1}&{\beta_t^2 - {{[\beta_t^2]}^2}}&{ - \beta_t^2\beta_t^3}& \cdots &{ - \beta_t^2\beta_t^M}\\
 \vdots & \vdots & \vdots & \cdots & \vdots \\
{ - \beta_t^M\beta_t^1}&{ - \beta_t^M\beta_t^2}&{ - \beta_t^M\beta_t^3}& \cdots &{\beta_t^M - {{[\beta_t^M]}^2}}
\end{array}} \right]
\end{equation}
Let ${\mathcal{A}_{t}}$ denotes the rightmost matrix in Eq.(5) and ${{{\mathcal{A}_{{t}}}} \mathord{\left/
 {\vphantom {{{A_{{t}}}} {{t}}}} \right.
 \kern-\nulldelimiterspace} {{t}}}$ is the Jacobian with the temperature $t$. Let $l$ indicates the loss actually used, then the backpropagation to $a$ can be written as:
\begin{equation}
\frac{{\partial {{\beta}_t}}}{{\partial {A}}}\frac{{\partial l}}{{\partial {{\beta}_t}}} = \frac{\mathcal{A}_{{t}}}{t}\frac{{\partial l}}{{\partial {{\beta}_t}}}
\end{equation}
By leveraging a smaller temperature $t$ to get a sparser $\beta_t$, only one single entry in $\beta_t$ approaches one (${\beta^i_t} \to 1$) and the others are thereby close to zero (${\beta^{j \ne i}_t} \to 0$), Jacobian matrix is then overall trapped in zero $\mathop {\lim }\limits_{\scriptstyle{\beta^i_t} \to 1\hfill\atop
\scriptstyle{\beta^{j \ne i}_t} \to 0\hfill} {{\partial {{\beta}_t}} \mathord{\left/
 {\vphantom {{\partial {{\beta}_t}} {\partial {A}}}} \right.
 \kern-\nulldelimiterspace} {\partial {A}}} = 0$ and the gradient saturates. When softmax is saturated, back propagation depicted in Eq.(6) stops propagating gradients through softmax to parameters thus the parameters stop updating. To deal with this problem, ideally, we hope that our softmax outputs a sparse $\beta$ in feedforward as the middle pane of Figure~\ref{fig1}, but the backpropagation is smoother (not saturated). To this end, we propose \textbf{s}parse-\textbf{n}oisy softmax (sn-softmax), summarized in Algorithm 1, to approximate the ideal case by \textit{combining different temperature for feedforward and backpropagation} respectively. Our goal is to keep the Jacobian matrix not zero even softmax converges $\mathop {\lim }\limits_{\scriptstyle{\beta^i_t} \to 1\hfill\atop
\scriptstyle{\beta^{j \ne i}_t} \to 0\hfill} {{\partial {{\beta}_t}} \mathord{\left/
 {\vphantom {{\partial {{\beta}_t}} {\partial {A}}}} \right.
 \kern-\nulldelimiterspace} {\partial {A}}} \ne 0$, which is different to the previous research that injects noise to postpone convergence~\cite{chen2017noisy}. 

 Set the forward-pass temperature as $t$, we formulate the Jacobian matrix of sn-softmax as:
\begin{equation}
\begin{aligned}
\frac{{\partial {{\beta}_{{t}}}}}{{\partial {A}}} = \frac{{{\mathcal{A}_{{t}}}}}{{{t}}} + \frac{{{\mathcal{A}_{{s}{t}}}}}{{{s}{t}}}{\rm{\ \ for}}\ s > 1
\end{aligned}
\end{equation}
  \begin{wrapfigure}{R}{0.55\textwidth}
     \vskip -0.54in
    \begin{minipage}{0.55\textwidth}
\begin{algorithm}[H]  
   \caption{Sparse-noisy (sn) softmax}
   \label{alg:1}
\begin{algorithmic}
   \STATE {\bfseries Input:} logits $A= \{ {a^1},...,{a^M}\}$, feedforward temperature $t$, scaling factor $s$.
   \STATE {\bfseries Output:} normalized distribution ${\beta_{{t}}}$
   \STATE {\bfseries Feedforward:} ${\rm softmax} (\frac{A}{{{t}}}) = {\beta_{{t}}}\ {\rm{for}}\ \beta_t^{{i}} = \frac{{\exp ({{{a^i}} \mathord{\left/
 {\vphantom {{{a^i}} {{t}}}} \right.
 \kern-\nulldelimiterspace} {{t}}})}}{{\sum\nolimits_{j}^M {\exp ({{{a^j}} \mathord{\left/
 {\vphantom {{{a^i}} {{t}}}} \right.
 \kern-\nulldelimiterspace} {{t}}})} }}$
   \STATE {\bfseries Backpropagation:} $\frac{{\partial {{\beta}_{{t}}}}}{{\partial {A}}} = \frac{1}{{{t}}}{\mathcal{A}_{{t}}} + \frac{1}{{{s}{t}}}{\mathcal{A}_{{s}{t}}}$ for ${s} > 1$
\end{algorithmic}
\end{algorithm}
    \end{minipage}
    \vskip -0.3in
  \end{wrapfigure}
where ${{{\mathcal{A}_{{s}{t}}}} \mathord{\left/
 {\vphantom {{{A_{{t}}}} {{t}}}} \right.
 \kern-\nulldelimiterspace} {{s}{t}}}$ is the Jacobian of $\beta_{{s}{t}}$. The backpropagation from the loss $l$ w.r.t $A$ can be written as:
\begin{equation}
\begin{aligned}
\frac{{\partial {{\beta}_{{t}}}}}{{\partial {A}}}\frac{{\partial l}}{{\partial {{\beta}_{{t}}}}} = (\frac{{{\mathcal{A}_{{t}}}}}{{{t}}} + \frac{{{\mathcal{A}_{{s}{t}}}}}{{{s}{t}}})\frac{{\partial l}}{{\partial {{\beta}_{{t}}}}}
\end{aligned}
\end{equation}
Since the equal sign can not be guaranteed:
\begin{equation}
sign(\frac{{{\mathcal{A}_{{t}}}}}{{{t}}}\frac{{\partial l}}{{\partial {{\beta}_{{t}}}}}) \not\equiv sign(\frac{{{\mathcal{A}_{s{t}}}}}{{s{t}}}\frac{{\partial l}}{{\partial {{\beta}_{{t}}}}})\ \ {\rm{for}}\ s > 1
\end{equation}
Sn-softmax is equivalent to adding noise into the backpropagation by which the Jacobian matrix is not absolute zero after the forward term ${{{\mathcal{A}_{{t}}}} \mathord{\left/
 {\vphantom {{{\mathcal{A}_{{t}}}} {{t}}}} \right.
 \kern-\nulldelimiterspace} {{t}}}$ converges. $s$ acts as a scaling factor in Eq.(8) to tune the noise intensity. A smaller $s$ brings stronger gradient noise, but is easier to saturate either as the training progresses. We can also keep $s{t}$ as a constant and determine $s$ accordingly and dynamically and thereby get out of the saturation in the whole training phase.

\begin{wrapfigure}[]{r}{0.5\textwidth}
\begin{center}
\vskip -0.48in
\includegraphics[width=.5\textwidth]{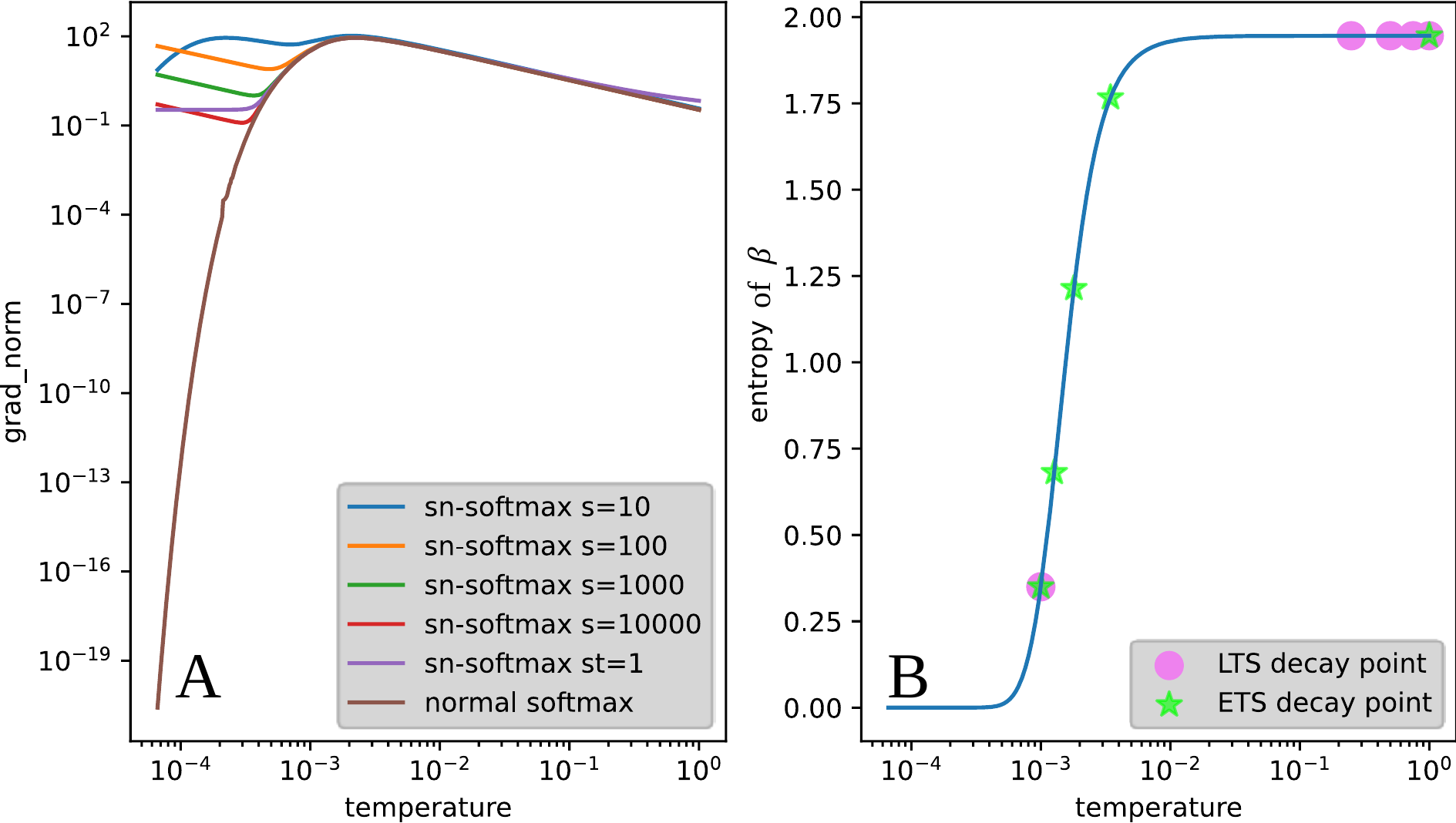}
\caption{Visualize the empirical analysis of sn-softmax and ETS. Tests are based on the order of magnitude of $a$ initialized in DARTS. (A). By setting $s \gg 1$, gradients of sn-softmax is consistent with normal softmax before saturated. After saturated, we can get different backpropagation dynamics by setting different $s$. (B). For equidistant decay, ETS yields much smoother swings than the linear counterpart in terms of the entropy of outbound distribution $\beta$.}
\label{sn-exp}
\end{center}
\vskip -0.33in
\end{wrapfigure}

In practice, sn-softmax leverages a small temperature value $t$ for feedforward to obtain a sparse output ${\beta_{{t}}}$ while setting $s \gg 1$ (generally $s>50$) thus get ${1 \mathord{\left/
 {\vphantom {1 {s{t}}}} \right.
 \kern-\nulldelimiterspace} {s{t}}} \ll {1 \mathord{\left/
 {\vphantom {1 {{t}}}} \right.
 \kern-\nulldelimiterspace} {{t}}}$. When the feedforward does not converge (${\mathcal{A}_{{t}}} \ne 0$), ${{{\mathcal{A}_{{t}}}} \mathord{\left/
 {\vphantom {{{\mathcal{A}_{{t}}}} {{t}}}} \right.
 \kern-\nulldelimiterspace} {{t}}}$ plays a leading role in Eq.(7) because of ${1 \mathord{\left/
 {\vphantom {1 {s{t}}}} \right.
 \kern-\nulldelimiterspace} {s{t}}} \ll {1 \mathord{\left/
 {\vphantom {1 {{t}}}} \right.
 \kern-\nulldelimiterspace} {{t}}}$ and 
 ${{{\mathcal{A}_{s{t}}}} \mathord{\left/
 {\vphantom {{{\mathcal{A}_{s{t}}}} {s{t}}}} \right.
 \kern-\nulldelimiterspace} {s{t}}} \ll {{{\mathcal{A}_{{t}}}} \mathord{\left/
 {\vphantom {{{\mathcal{A}_{{t}}}} {{t}}}} \right.
 \kern-\nulldelimiterspace} {{t}}}$. After the feedforward convergence, the gradient saturates (${\mathcal{A}_{{t}}} = 0$, ${{{\mathcal{A}_{{t}}}} \mathord{\left/
 {\vphantom {{{\mathcal{A}_{{t}}}} {{t}}}} \right.
 \kern-\nulldelimiterspace} {{t}}} = 0$). Since the scaling factor $s \gg 1$, ${\beta_{s{t}}}$ is much smoother than ${\beta_{{t}}}$ which makes ${\mathcal{A}_{s{t}}}$ less convergent (${\mathcal{A}_{s{t}}} \ne 0$). As a result, ${{{\mathcal{A}_{s{t}}}} \mathord{\left/
 {\vphantom {{{\mathcal{A}_{s{t}}}} {s{t}}}} \right.
 \kern-\nulldelimiterspace} {s{t}}}$ supersedes the first term ${{{\mathcal{A}_{{t}}}} \mathord{\left/
 {\vphantom {{{\mathcal{A}_{{t}}}} {{t}}}} \right.
 \kern-\nulldelimiterspace} {{t}}}$ in the Jacobian matrix and comes into play in the backpropagation of sn-softmax depicted in Eq.(8). This way, when sn-softmax does not converge, the normal term repels the noise term and dominate the update direction due to  ${{{\mathcal{A}_{s{t}}}} \mathord{\left/
 {\vphantom {{{\mathcal{A}_{s{t}}}} {s{t}}}} \right.
 \kern-\nulldelimiterspace} {s{t}}} \ll {{{\mathcal{A}_{{t}}}} \mathord{\left/
 {\vphantom {{{A_{{t}}}} {{t}}}} \right.
 \kern-\nulldelimiterspace} {{t}}}$ but allows sn-softmax to have a chance to escape the premature convergence after it's saturated. We schematically visualize the backpropagation dynamic of sn-softmax in terms of the gradient norm on different values of $s$ in Figure~\ref{sn-exp}A.
 
 Sn-softmax only needs \textit{feedforward and backpropagation once like softmax} to get ${{\partial l} \mathord{\left/
 {\vphantom {{\partial l} {\partial {{\beta}_{{t}}}}}} \right.
 \kern-\nulldelimiterspace} {\partial {{\beta}_{{t}}}}}$. Then it calculates the smoother Jacobian based on temperature $s{t}$ through Eq.(5). The calculation of another Jacobian and the final addition depicted in Eq.(7) only takes negligible additional budget in the whole training.

\subsection{Exponential Temperature Schedule}
In general, the temperature $t$ is gradually decayed to drive $\beta$ converge to a sparse solution. As shown in Eq.(3), $t$ is in the exponential term and nonlinearly affects the transformation of $a\to\beta$ in softmax. Linearly scheduling $t$ swings $\beta$ nonlinearly and results in that the early temperature decay has little effect on the $\beta$ while decay in the later stage of training has too much impact on the sparsity and causes the training converges precipitately (see LTS in Figure~\ref{sn-exp}B and LPCD-DARTS in Figure~\ref{fig4}A). Therefore, the naive linear temperature schedule is inappropriate for the training of DARTS.
In this section, we focus on temperature $t$ so let $t_a^{\exp } = \exp ({a \mathord{\left/
 {\vphantom {a t}} \right.
 \kern-\nulldelimiterspace} t})$ refers to the exponential function with $e$ as the base and $t$, $a$ as the exponent where we add the superscript “exp” in $t_a^{\exp }$ to emphasize the value after the exponential transformation $\exp ({a \mathord{\left/
 {\vphantom {a t}} \right.
 \kern-\nulldelimiterspace} t})$. We then apply variable substitution in softmax as:
\begin{equation}
{\beta^i} = \frac{{t_{{a^i}}^{\exp }}}{{\sum\nolimits_{j = 1}^M {t_{{a^j}}^{\exp }} }}\ \ {\rm{where}}\ t_{a^\ast}^{\exp}=\exp(\frac{a^\ast}{t})
\end{equation}
where the transformation from $t_{{a}}^{\exp }$ to ${\beta^i}$ is linear. We can then linearly schedule $t_{{a}}^{\exp }$ instead of $t$ to better control the sparsity of $\beta$. We call this design of scheduling $t_{{a}}^{\exp }$ after the exponent substitution as exponential temperature schedule (ETS).

To calculate $t_{{a}}^{\exp }$, we first approximate $a$ by the order of its expectation $E\left( a \right)$, $a\in A$ and get the temperature $t_{{E(a)}}^{\exp }$. From now on, we omit $E(a)$ in the subscript of $t_{{E(a)}}^{\exp }$ since it's fixed in this analysis. 
We then specify the initial temperature ${t_{0}}$ and the decay target ${t_{N}}$ and get $t_{0}^{\exp } = {e^{{{E(a)} \mathord{\left/
 {\vphantom {{E(a)} {{t_{0}}}}} \right.
 \kern-\nulldelimiterspace} {{t_{0}}}}}}$ and $t_{N}^{\exp } = {e^{{{E(a)} \mathord{\left/
 {\vphantom {{E(x)} {{t_{N}}}}} \right.
 \kern-\nulldelimiterspace} {{t_{N}}}}}}$ for the start and target in the exponential space (after $\exp ({a \mathord{\left/
 {\vphantom {a t}} \right.
 \kern-\nulldelimiterspace} t})$ transformation) respectively. Since $t$ act as the denominator in the exponential term $\exp ({a \mathord{\left/
 {\vphantom {a t}} \right.
 \kern-\nulldelimiterspace} t})$ of softmax, the temperature decay from ${t_{0}}$ to ${t_{N}}$ corresponds to an ascent in the exponential space from $t_{0}^{\exp }$ to $t_{N}^{\exp }$. For the equidistant temperature decay, the decay strength $d^{\exp }$ indicates the variation amplitude within $[t_{0}^{\exp },t_{N}^{\exp }]$ which can be calculated statically via $d^{\exp }=\frac{{t_{N}^{\exp } - t_{0}^{\exp }}}{N}$ where ${N}$ is the number of decay points. All decay points together with $t_{0}^{\exp }$ form a list ${L^{\exp }}$ as
\begin{equation}
\begin{aligned}
{L^{\exp }} = [t_n^{\exp }|t_n^{\exp } = t_{0}^{\exp } + nd^{\exp },\ d^{\exp }= 
\frac{{t_{N}^{\exp } - t_{0}^{\exp }}}{{{N}}},n = 0,1,...,{N}]
\end{aligned}
\end{equation}
where $t_0^{\exp } = t_{0}^{\exp }$, $t_{{N}}^{\exp } = t_{N}^{\exp }$. The temperature value corresponding to each decay point in ${L^{\exp }}$ can be inversely solved by Eq.(12) where $t_{n} > 0$ when ${t_{n}^{\exp }} > 1$. 
\begin{equation}
{t_{n}} = \frac{{E(a)}}{{\ln ({t_{n}^{\exp }})}}\ \ {\rm{for}}\ {t_{n}^{\exp }} \in {L^{\exp }},\ n = 0,1,...,{N}
\end{equation}

\begin{wrapfigure}[]{r}{0.3\textwidth}
\begin{center}
\vskip -0.46in
\includegraphics[width=.3\textwidth]{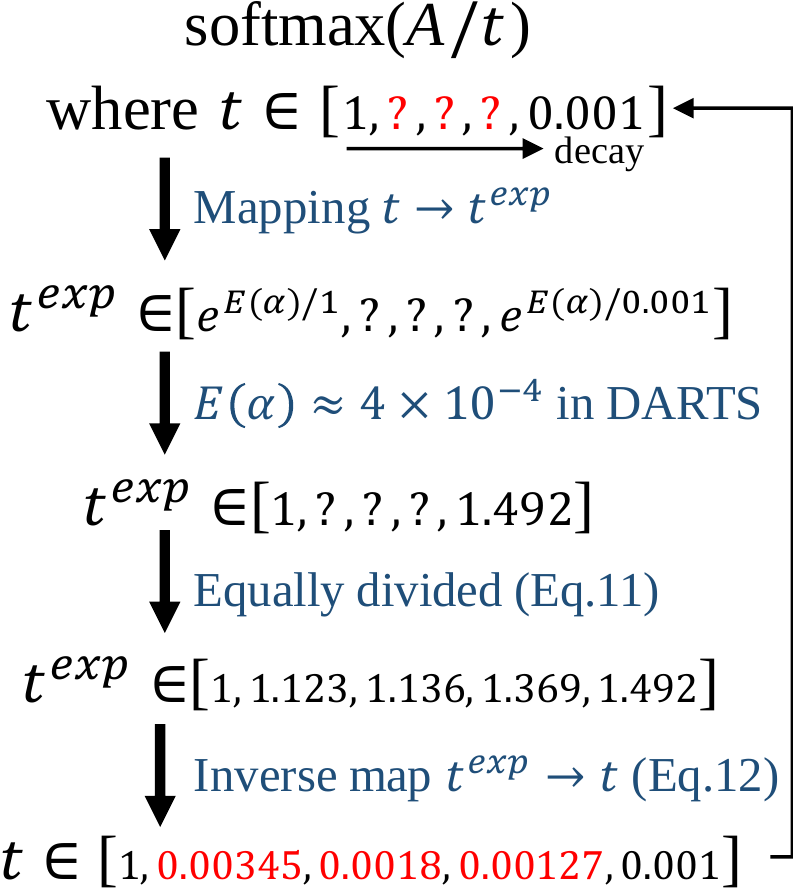}
\caption{Diagram of the quantitative example.}
\label{ets-digram}
\end{center}
\vskip -0.33in
\end{wrapfigure}

We provide a quantitative example in Figure~\ref{ets-digram} to show the clear usage of ETS and more details on the calculation are given below. Firstly, $a$s in $A$ are initialized by sampling from $N(0,1)$ and scaling the samples by $1$$e$-$3$ in DARTS. 
When $a \leq 0$, the variation of $t$ has little effect on the value of $t_{{a}}^{\exp }$, so we only consider $E(a)$ for ${a} > 0$. The scaled expectation of ${a}$ can be get through $ E(a)$ = $(1e{\rm{-}}3)E(N(0,1))$=$(1e{\rm{-}}3)\sqrt {2/\pi } /2
 \approx 4e{\rm{-}}4$ for ${a} > 0$. By specifying the initial temperature ${t_{0}} = 1$ and the decay target ${t_{N}} = 1e$-$3$, we can then get $t_{0}^{\exp }$= $\exp(E(a)/t_{0})$=${e^{4e {\rm{-}} 4}} \approx 1$ and $t_{N}^{\exp }$ = $\exp(E(a)/t_{N})$=${e^{4e {\rm{-}} 1}} \approx 1.492$ in the exponential space respectively. The temperature decay from ${t_{0}}$ to ${t_{N}}$ ($1 \to 1e$-$3$) thus corresponds to an ascent from $t_{0}^{\exp }$ to $t_{N}^{\exp }$ ($1 \to 1.492$) in the exponential space. For the equidistant temperature decay, we preset ${N} = 4$ and  $t_4^{\exp } = t_{N}^{\exp } = 1.492$, then $[1,1.492]$ is equidistantly divided into 4 segments by the other 3 decay points. We get the decay strength accordingly as $d^{\exp } = \frac{{1.492 - 1}}{4} = 0.123$ and determine the remaining three decay points as $1.123$, $1.246$, $1.369$ and form ${L^{\exp }} = [1,1.123,1.246,1.369,1.492]$ by Eq.(11). After that, we can get the temperatures w.r.t $t_{1.123}^{\exp }$ as ${t_{1.123}} = {{4e {\rm{-}} 4} \mathord{\left/
 {\vphantom {{4e - 4} {\ln (1.123)}}} \right.
 \kern-\nulldelimiterspace} {\ln (1.123)}} \approx 0.00345$ by Eq.(12). Similarly, the temperature of the other 3 decay points are ${t_{1.246}} = 0.0018$, ${t_{1.369}} = 0.00127$, ${t_{1.492}} = 0.001$ from which we can clearly find that the temperature decay scheduled equidistantly in the exponential space ($1,0.00345,0.0018,0.00127,0.001$) leads to a radical difference from the typical linear temperature scheduling LTS ($1,0.75,0.5,0.25,0.001$). We visualize the effect of the differences between ETS and LTS in terms of the entropy of $\beta$ in a more sensible way in Figure~\ref{sn-exp}B.
 
For linear temperature schedule, we always need to preset ${t_{0}}$, ${t_{N}}$, ${N}$ and calculate the decay strength statically to \textit{prevent temperature from decreasing to less than $0$}. In contrast, Eq.(12) is always greater than $0$ for ${t^{\exp }} > 1$ which yields added flexibility for the design of ETS-based training scheme. We can calculate $d^{\exp }$ and ${L^{\exp }}$ after we specified $[t_{0}^{\exp },t_{N}^{\exp }]$ and ${N}$. Alternatively, we can first determine $t_{0}^{\exp }$, $d^{\exp }$ and get $t_{N}^{\exp }$ or ${N}$ by $t_{N}^{\exp } = t_{0}^{\exp } + {N}d^{\exp }$, so that we can still build ${L^{\exp }}$ through Eq.(11).

\subsection{Entropy-based Adaptive Scheme}
  \begin{wrapfigure}{R}{0.55\textwidth}
     \vskip -0.6in
    \begin{minipage}{0.55\textwidth}
\begin{algorithm}[H]  
   \caption{EDD sparse training scheme}
   \label{alg:2}
\begin{algorithmic}
\STATE {\bfseries Input:} $s$ for sn-softmax, $\lambda $ for EDD, $t_{0}=1$ (mostly)
\STATE Get the expectation of parameter $E(a)$
\STATE Get $t_{0}^{\exp } = {e^{{{E(a)} \mathord{\left/
 {\vphantom {{E(x)} {{t_{0}}}}} \right.
 \kern-\nulldelimiterspace} {{t_{0}}}}}}$
\WHILE{training epoch $k$}
\STATE Training $a$ and $\omega$ by Algorithm 1 in \cite{liu2018darts} with ${t^{(k)}}$, $s$.
\STATE Update $ {[d^{\exp }]^{(k)}} $ by Eq.(14).
\STATE Update ${t^{(k)}}$ via $ {[d^{\exp }]^{(k)}} $ by Eq.(15).
\ENDWHILE
\end{algorithmic}
\end{algorithm}
    \end{minipage}
    \vskip -0.25in
  \end{wrapfigure}
If operation output $o_{c}^{i}(h)$ on edge $c$ is zero-weighted by as $\beta_{c}^{i}o^{i}_c(h)$ for $\beta_{c}^{i}\in\beta_{c}$ and $\beta_{c}^{i}=0$, both forward and backward paths of the operation $o^{i}_c$ are blocked so that the operation weights $\omega$ within $o_{c}^{i}$ cannot obtain effective gradients. Employing a small temperature at the beginning of training will lead to exaggerated swings of $\beta$ and finally biases the search result.
For an appropriate scheme, presetting fixed ${t_{0}}$, ${t_{N}}$, $d^{\exp }$ and updating $E\left( x \right)$ and ${L^{\exp }}$ every epoch is \textit{cumbersome and inflexible}. Empirically, we also observe that the same architecture under different initialization exhibits various optimization dynamics. Some lead to strong convergence that the parameters of supernet converge quickly under mild temperature decay. In other cases, the search is indecisive among two or three operations. In sum, a fixed decay strength $d^{\exp }$ is \textit{not robust in practice}. 

Being equipped with the additional design freedom supported by ETS, we further propose \textbf{e}ntropy-based \textbf{d}ynamic \textbf{d}ecay (EDD) to adaptively determine $d^{\exp }$ in terms of both the sparsity of $\beta$ and the current epoch $k$ by which \textit{we need only to tune one single hyperparameter $\lambda$ to control the training process}. 
We first define the expectation of entropies of $\beta$s as $E(H(\beta_c))$, $c\in C$ in Eq.(13) to represent the sparsity of $\beta$s over all compound edges in the cell space.
\begin{equation}
\begin{aligned}
{E(H(\beta_c))} = \frac{{ -\sum\nolimits_{c}^{|C|} {\sum\nolimits_i^M {\beta_c^i\log \beta_c^i} } }}{|C|}\ \ {\rm{where}}\ {\beta_{c}} = {\rm{softmax}} ({A_{c}}),\ c\in C
\end{aligned}
\end{equation}
where $M$ operation candidates and $|C|$ compound edges in search space. $\beta_c^i$ denotes the $i$th entry in the $\beta_{c}$ that used to weight the feature maps from operation $o^{i}$ on edge $c$ . The design principles can be summarized as follows:
\begin{itemize}
\setlength{\itemsep}{0pt}
\setlength{\parsep}{0pt}
\setlength{\parskip}{0pt}
\item $d^{\exp }$ is stronger when $E(H(\beta))$ is higher (smoother $\beta$ in softmax);
\item Gradually increase $d^{\exp }$ w.r.t the training epoch $k$.
\end{itemize}

Based on the above design principles, We update $d^{\exp }$ according to Eq.(14) for epoch $k$.
\begin{equation}
{[d^{\exp }]^{(k)}} = \lambda(1 - \rho )E(H(\beta))  + \rho {[d^{\exp }]^{(k - 1)}}
\end{equation}
where we set ${[d^{\exp }]^{(0)}}=0$ in practice and keep an exponentially moving average with a momentum $\rho \equiv 0.5$ to avoid oscillations. $\lambda $ is the hyperparameter to determine the influence of $E(H(\beta))$ on $d^{\exp }$. Since $d^{\exp }$ is adaptively decided through Eq.(14), ${L^{\exp }}$ cannot be calculated in advance. After $k$th update of $d^{\exp }$, we can get temperature ${t^{(k)}}$ accordingly by Eq.(15)
\begin{equation}
{t^{(k)}} = \frac{{{E(a)}}}{{\ln(t_{0}^{\exp } + k{{[d^{\exp }]}^{(k)}})}}
\end{equation}
results in that $d^{\exp }$ is proportional to the training epoch $k$. EDD is summarised in Algorithm 2 where $\lambda $ is the only hyperparameter for the scheme.

\begin{table*}[t]
\caption{Evaluation results on NB201\&C10. “clip” refers to the gradient clip on $a$ to alleviate the effect of the noisy none (zero) operation which has been identified as unsearchable in DARTS~\cite{liu2018darts}.
}
\vskip -0.18in
\label{table1}
\begin{center}
\resizebox{\textwidth}{!}{
\begin{tabular}{ccccccccc}
\toprule
\multirow{2}{*}{\begin{tabular}[c]{@{}c@{}}Search\\ dataset\end{tabular}}&\multirow{2}{*}{Method} & \multirow{2}{*}{\begin{tabular}[c]{@{}c@{}}Search\\ (seconds)\end{tabular}} & \multicolumn{2}{c}{CIFAR-10} & \multicolumn{2}{c}{CIFAR-100} & \multicolumn{2}{c}{ImageNet-16-120} \\
& &  & validation & test & validation & test & validation & test \\
\midrule
\multirow{9}{*}{C10}&DARTS-V2 \cite{liu2018darts} & 22323 & 39.77±0.00 & 54.30±0.00 & 15.03±0.00 & 15.61±0.00 & 16.43±0.00 & 16.32±0.00 \\
&DARTS-V1 \cite{liu2018darts} & 7253 & 39.77±0.00 & 54.30±0.00 & 15.03±0.00 & 15.61±0.00 & 16.43±0.00 & 16.32±0.00 \\
&GDAS \cite{dong2019searching} & 19720 & 90.00±0.21 & 93.51±0.13 & 71.14±0.27 & 70.61±0.26 & 41.70±1.26 & 41.84±0.90 \\
&\textbf{ETS-GDAS} & 19755 & 90.57±0.35 & 93.75±0.22 & 71.45±0.64 & 71.39±0.66 & 42.96±0.93 & 42.92±0.87\\
&DrNAS \cite{chen2021drnas} & 7544 & 90.15±0.10 & 93.74±0.03 & 70.82±0.27 & 71.07±0.08 & 40.76±0.05 & 41.37±0.17 \\
&GAEA-Bilevel \cite{li2021geometry} & 8280 & 82.80±1.01 & 84.64±1.00 & 55.24±1.47 & 55.35±1.72 & 27.72±1.35 & 26.40±0.85 \\
&GAEA-ERM \cite{li2021geometry} & 14464 & 84.59±0.00 & 86.59±0.00 & 58.12±0.00 & 58.43±0.00 & 29.54±0.00 & 28.19±0.00 \\
& &  & (91.50±0.06) & (94.34±0.06) & (73.12±0.26) & (73.11±0.06) & (45.71±0.28) & (46.38±0.18) \\
&GibbsNAS \cite{xue2021rethinking} & - & 90.02±0.60 & 92.72±0.60 & 68.88±1.43 & 69.20±1.44 & 42.31±1.69 & 42.08±1.95 \\
&DARTS- \cite{chu2020darts} & - & 91.03±0.44 &93.80±0.40 &71.36±1.51 &71.53±1.51 &44.87±1.46 &45.12±0.82\\
&SurgeNAS \cite{luo2022surgenas} & - & 90.2 &93.7 &71.2 &71.6 &44.5 &45.2\\
&\textbf{EDD-DARTS} & 7392 & 90.95±0.44 & \textbf{93.80±0.33} & \textbf{71.44±1.21} & 71.42±1.25 & \textbf{45.14±0.78} & 45.12±0.56 \\
&\textbf{EDD-DARTS (clip)}& 7400 & \textbf{91.12±0.19} & \textbf{94.05±0.11} & \textbf{72.59±0.62} & \textbf{72.43±0.64} & \textbf{45.89±0.41} & \textbf{45.80±0.29} \\
\bottomrule
\end{tabular}
}
\end{center}
\caption{Experimental results on S1$\sim$S4\&C100 and S1$\sim$S4\&SVHN.}
\vskip -0.3in
\label{table6}
\begin{center}
\resizebox{\textwidth}{!}{
\begin{tabular}{ccccccccc}
\toprule
Dataset &Space & DARTS & PC-DARTS & DARTS-ES & R-DARTS (DP/L2) & SDARTS (RS/ADV) & \begin{tabular}[c]{@{}c@{}}DARTS+PT\\    (unfixed/fixed) \end{tabular} & \textbf{EDD-DARTS}\\
\midrule
\multirow{4}{*}{C100} &S1 & 29.46 & 24.69 & 28.37 & 25.93/24.25 & 23.51/22.33 & 24.48/24.40 & \textbf{22.27}\\
&S2 & 26.05 & 22.48 & 23.25 & 22.30/22.44 & 22.28/\textbf{20.56} & 23.16/23.30 & 21.73\\
&S3 & 28.90 & 21.69 & 23.73 & 22.36/23.99 & 21.09/\textbf{21.08} & 22.03/21.94 & \textbf{21.08} \\
&S4 & 22.85 & 21.50 & 21.26 & 22.18/21.94 & 21.46/21.25 & 20.80/\textbf{20.66} & \textbf{20.66}\\
\midrule
\multirow{4}{*}{SVHN} &S1 & 4.58 & 2.47 & 2.72 & 2.55/4.79 & 2.35/2.29 & 2.62/2.39 & \textbf{2.23}\\ 
&S2 & 3.53 & 2.42 & 2.60 & 2.52/2.51 & 2.39/2.35 & 2.53/2.32 & \textbf{2.30}\\ 
&S3 & 3.41 & 2.41 & 2.50 & 2.49/2.48 & 2.36/2.40 & 2.42/\textbf{2.32} & \textbf{2.32}\\ 
&S4 & 3.05 & 2.43 & 2.51 & 2.61/2.50 & 2.46/2.42 & 2.42/\textbf{2.39} & 2.45\\
\bottomrule
\end{tabular}
}
\end{center}
\vskip -0.3in
\end{table*}

\section{Evaluations}
We evaluate EDD, namely EDD-DARTS, on CIFAR-10 (C10), CIFAR-100 (C100), ImageNet-1k (IN-1k), SVHN and multiple spaces: NAS-BENCH-201 (NB201), DARTS space (DS), S1$\sim$S4 \cite{arber2020understanding}.

\textbf{Evaluations on NB201}: 
NB201 supports three datasets (C10, C100, ImageNet-16~\cite{chrabaszcz2017downsampled}) and has a unified cell-based search space with 15,625 architectures. We refer their paper \cite{dong2019bench} for more details of the search space. All our baselines come from recent top venues. Experiments of DrNAS and GAEA are both based on the released codes. We provide extra results for GAEA-ERM in parentheses by excluding the none operation since it's particularly fragile for "none" in NB201. The experimental results shown in Table~\ref{table1} demonstrates that the our enhancement effectively eliminates the performance collapse of DARTS. Remarkably, EDD-DARTS claims superior searching on both C10 and C100 on the  standard space (include none) of NB201.

\textbf{Evaluations on DS}: We employ the same search recipe as on NB201. Our evaluation is based on the source code released by DrNAS. We keep the hyperparameter settings unchanged except for replacing the cell genotypes. As~\cite{chen2021drnas,li2021geometry}, we repeat the search 3 times under different seeds, evaluate each result independently and report the mean accuracies and standard deviations in Table~\ref{table4}.

\begin{wraptable}[]{r}{0.49\textwidth}
\vskip -0.45in
\caption{Search and evaluate on DS\&C10.}
\label{table4}
\begin{center}
\resizebox{0.49\textwidth}{!}{
\begin{tabular}{cccc}
\toprule
Method & Error (\%) &Params (M) &  GPU days\\
\midrule
PC-DARTS \cite{xu2019pc} &     2.57±0.07 &   3.6& 0.1 \\
GAEA+PC-DARTS \cite{li2021geometry}        &     2.50±0.06 &   3.7      & 0.1 \\
DARTS+PT \cite{wang2020rethinking}         &    2.61±0.08&  3.0     & 0.8  \\
SDARTS-RS+PT \cite{wang2020rethinking}           &    2.54±0.10    &  3.3  & 0.8   \\
SGAS+PT \cite{wang2020rethinking}       &     2.56±0.10   &    3.9    & 0.29   \\
DrNAS \cite{chen2021drnas}      &    2.54±0.03  &    4.0  &0.4      \\
DARTS- \cite{chu2020darts} & 2.59±0.08&3.5&0.4\\
GibbsNAS \cite{xue2021rethinking} & 2.53±0.02&4.1&0.5\\
SparseNAS \cite{wu2021neural}& 2.69±0.03&4.2&1\\
$\beta$-DARTS \cite{ye2022b} & 2.53±0.08&3.7&0.4\\
\midrule
EDD-DARTS        &     2.52±0.10&     3.6    &0.4     \\
EDD-PC-DARTS         &     \textbf{2.47±0.06}  &     4.2  &\textbf{0.1}    \\
\bottomrule
\end{tabular}
}
\end{center}
\vskip -0.15in
\caption{Transfer to evaluate on IN1K.}
\label{table5}
\begin{center}
\resizebox{0.47\textwidth}{!}{
\begin{threeparttable}
\begin{tabular}{ccccc}
\toprule
Method                  & Top-1 &Top-5& Params (M) & GPU days \\
\midrule
PC-DARTS  \cite{xu2019pc}         & 25.1       & 7.8        & 5.3        & 0.1                    \\
GAEA+PC-DARTS  \cite{li2021geometry}     & 24.3       & 7.3        & 5.6        & 0.1                    \\
GibbsNAS \cite{xue2021rethinking} &24.6&-&5.1&0.5 \\
DrNAS \cite{chen2021drnas} &24.2&7.3&5.2&3.9\tnote{*} \\
DARTS- \cite{chu2020darts} &24.8& \textbf{7}&4.9&4.5\tnote{*}\\
SparseNAS \cite{wu2021neural}& 24.6&7.6&5.7&-\\
$\beta$-DARTS \cite{ye2022b} & 24.2&7.1&5.4&0.4\\
\midrule
EDD-DARTS & 24.6       & 7.4        & 5.0        & 0.4                    \\
\textbf{EDD-PC-DARTS}  & \textbf{24.0}       & 7.2        & 5.6        & \textbf{0.1} \\
\bottomrule
\end{tabular}
\begin{tablenotes}\footnotesize
\item[*]Search on ImagetNet.
\end{tablenotes}
 \end{threeparttable}
}
\end{center}
\vskip -0.35in
\end{wraptable}

\textbf{Transfer to ImageNet-1k}: As a common practice, we transfer the most prominent architecture on C10 to ImageNet-1k (IN1K) for additional performance evaluation. As shown in Table~\ref{table4} and Table~\ref{table5}, the EDD itself is already very competitive, and the combination of the EDD with partial channel trick~\cite{xu2019pc} can deliver a clear new art scores with the lowest budget on both C10 and ImageNet-1k.

\textbf{Comparison with Regularization}: 
We evaluate EDD on four specially designed search spaces S1$\sim$S4 by \cite{arber2020understanding} which are particularly challenging for DARTS-based methods. Unregularized DARTS is always prone to make a wrong choice of non-parametric operations on these spaces. We refer to \cite{arber2020understanding} for more details of S1$\sim$S4. We evaluate our method against two strong baselines Smooth DARTS \cite{chen2020stabilizing} and DARTS+PT \cite{wang2020rethinking} on this benchmark on C10, C100 and SVHN respectively. 
As illustrated in Table~\ref{table6}, EDD-DARTS performs well under all datasets and search spaces and outperforms baselines in aggregate.

\section{Further Experiments, Analyses and Conclusion}

To understand what makes EDD effective, we conduct ablation studies along two axes, $s$ and $\lambda$, on both NB201 and DS. We come up with and elaborate another ETS-based baseline to validate our claim of superior robustness of EDD. 

\textbf{Ablations on NB201}: We ablate the impact of three configurations of softmax and $\lambda$ on NB201. The experimental results on C10 are shown in Table~\ref{table7}. For the searching on C10, sn-softmax is helpful where both the highest and second-highest accuracies in the experiments of EDD-DARTS come from the results equipped with sn-softmax ($s{t} \equiv 1$ or $s = 100$).

\begin{table*}[t]
\caption{Ablate $s$ and $\lambda$ on NB201\&C10.}
\vskip -0.3in
\label{table7}
\begin{center}
\resizebox{\textwidth}{!}{
\begin{tabular}{cccccccccc}
\toprule 
\multirow{2}{*}{\begin{tabular}[c]{@{}c@{}}Search\\ dataset\end{tabular}}& \multirow{2}{*}{Softmax}        & \multirow{2}{*}{$s$} & \multirow{2}{*}{$\lambda$} & \multicolumn{2}{c}{CIFAR-10} & \multicolumn{2}{c}{CIFAR-100} & \multicolumn{2}{c}{ImageNet-16-120} \\ &                                 &                      &                            & valiation     & test         & valiation     & test          & valiation        & test             \\ \midrule \multirow{9}{*}{C10}                                                      & \multirow{3}{*}{Normal softmax} & \multirow{3}{*}{-}   & 0.06                       & 90.56±0.75    & 93.29±0.70   & 70.03±1.48    & 70.21±1.54    & 43.59±2.05       & 43.33±1.80       \\ &                                 &                      & 0.12                       & 90.74±0.31    & 93.60±0.29   & 70.46±0.84    & 70.45±0.96    & 44.17±0.92       & 44.34±1.23       \\ &                                 &                      & 0.24                       & 89.93±1.44    & 92.82±1.16   & 69.38±2.05    & 69.60±1.49    & 43.01±2.16       & 43.07±2.14       \\ \cline{2-10} & \multirow{6}{*}{sn-softmax}     & \multirow{3}{*}{$s{t} \equiv 1$}    & 0.06                       & \textbf{
90.95±0.44}    & \textbf{93.80±0.33}   & \textbf{71.44±1.21}    & \textbf{71.42±1.25}    & \textbf{45.14±0.78}       & \textbf{45.12±0.56}       \\ &                                 &                      & 0.12                       & 90.57±0.58    & 93.38±0.52   & 70.06±1.46    & 69.86±1.42    & 43.87±1.92       & 43.89±2.33       \\ &                                 &                      & 0.24                       & 90.06±0.81    & 92.83±0.52   & 69.07±1.35    & 69.26±1.29    & 42.28±2.07       & 42.43±1.83       \\ \cline{3-10} &                                 & \multirow{3}{*}{$s = 100$}    & 0.06                       & 90.68±0.71    & 93.44±0.54   & 70.45±1.69    & 70.43±1.61    & 43.98±1.85       & 44.08±2.17       \\ &                                 &                      & 0.12                       & \textbf{91.02±0.30}    & \textbf{93.75±0.31}   & \textbf{71.36±0.96}    & \textbf{71.33±1.21}    & 44.73±0.64       & 44.95±0.62       \\ &                                 &                      & 0.24                       & 90.54±0.74    & 93.22±0.58   & 69.90±1.60    & 70.19±1.88    & 43.43±1.82       & 43.49±1.68       \\
\bottomrule
\end{tabular}
}
\end{center}
\vskip -0.4in
\end{table*}

\begin{wraptable}[]{r}{0.49\textwidth}
\vskip -0.42in
\caption{Ablate $s$ and $\lambda$ on DS\&C10.}
\vskip -0.2in
\label{table8}
\begin{center}
\resizebox{0.49\textwidth}{!}{
\begin{tabular}{ccccc}
\toprule
\multirow{2}{*}{\begin{tabular}[c]{@{}c@{}}Search\\ dataset\end{tabular}}&\multirow{2}{*}{$\lambda$}                   & normal softmax & sn-softmax $s{t} \equiv 1$ & sn-softmax $s=100$ \\
& &   Test Error (\%)    & Test Error (\%)                    & Test Error (\%)                           \\
\midrule
\multirow{3}{*}{C10}&0.06&         2.60 ±  0.10                  &       2.55  ±  0.10                                   &              \textbf{2.52  ±  0.10}                                      \\ &0.12                    &         2.63 ±  0.13                 &       2.56  ±  0.10                                   &              2.54  ±  0.09                           \\
&0.24                                     &        2.65 ±  0.13                     &         2.59 ±  0.12                 &                         2.55  ±  0.10              \\
\bottomrule
\end{tabular}
}
\end{center}
\vskip -0.4in
\end{wraptable}

\textbf{Ablations on DS}: We ablate EDD-DARTS further on DS and the results are shown in Table~\ref{table8}. We find that a larger $\lambda$ brings stronger temperature decay, EDD-DARTS tends to find higher capacity architectures some of which are tricky to be trained properly but sn-softmax alleviates this trend and ensures that EDD keeps delivering efficient results.

\begin{wraptable}[]{r}{0.52\textwidth}
\vskip -0.46in
\caption{Finetune (FT) on NB201\&C10 and transfer to other datasets and search spaces to evaluate the robustness.} \label{table9} 
\begin{center}
\resizebox{0.52\textwidth}{!}{ \begin{tabular}{c|cc|cc|cc} \toprule \multirow{2}{*}{\begin{tabular}[c]{@{}c@{}}Search\\ space\end{tabular}} & \multirow{2}{*}{\begin{tabular}[c]{@{}c@{}}Search\\ dataset\end{tabular}} & \multirow{2}{*}{\begin{tabular}[c]{@{}c@{}}Evaluation\\ dataset\end{tabular}} & \multicolumn{2}{c|}{PCD-DARTS}  & \multicolumn{2}{c}{EDD-DARTS}  \\                                                                             &                                                                                 &                                                                                     & validation      & test          & validation     & test          \\ \midrule \multirow{6}{*}{NB201}                                                          & \multirow{3}{*}{\begin{tabular}[c]{@{}c@{}}C10\\ (FT)\end{tabular}}                                                           & C10                                                                                 & 91.06±0.49      & 93.78±0.43    & 90.95±0.44     & 93.80±0.33    \\ &                                                                                 & C100                                                                                & 71.61±1.30      & 72.00±1.48    & 71.44±1.21     & 71.42±1.25    \\ &                                                                                 & IN-16                                                                                & 45.17±0.84      & 45.63±0.86   & 45.14±0.78     & 45.12±0.56    \\ \cline{2-7} & \multirow{3}{*}{C100}                                                           & C10                                                                                 & 90.04±1.35      & 92.97±1.78    & \textbf{90.27±0.81}     & \textbf{93.44±0.49}    \\ &                                                                                 & C100                                                                                & 69.97±1.82      & 70.02±2.00    & \textbf{70.46±1.52}     & \textbf{70.57±1.40}    \\ &                                                                                 & IN-16                                                                                & 42.34±2.16      & 42.79±2.01    & \textbf{42.43±1.80}     & \textbf{42.88±1.81}    \\ \midrule                                                                              &                                                                                 &                                                                                     & Error (\%)      & Param (M)     & Error (\%)     & Param (M)     \\ \midrule \multirow{2}{*}{DS}                                                          & \multirow{2}{*}{C10}                                                            & C10                                                                                 & 2.57±0.09       & 3.0           & \textbf{2.55±0.10}      & 3.6           \\ &                                                                                 & IN-1k                                                                               & 26.5            & 4.2           & \textbf{24.6}           & 5.1           \\ \midrule                                                                              &                                                                                 &                                                                                     & \multicolumn{2}{c|}{Error (\%)} & \multicolumn{2}{c}{Error (\%)} \\ \midrule \multirow{3}{*}{S1}                                                           & \multicolumn{2}{c|}{C10}                                                                                                                                              & \multicolumn{2}{c|}{\textbf{2.75}}       & \multicolumn{2}{c}{2.77}       \\ & \multicolumn{2}{c|}{C100}                                                                                                                                             & \multicolumn{2}{c|}{22.35}      & \multicolumn{2}{c}{\textbf{22.27}}      \\ & \multicolumn{2}{c|}{SVHN}                                                                                                                                             & \multicolumn{2}{c|}{2.27}       & \multicolumn{2}{c}{\textbf{2.23}}       \\ \midrule \multirow{3}{*}{S2}                                                           & \multicolumn{2}{c|}{C10}                                                                                                                                              & \multicolumn{2}{c|}{2.56}       & \multicolumn{2}{c}{\textbf{2.54}}       \\ & \multicolumn{2}{c|}{C100}                                                                                                                                             & \multicolumn{2}{c|}{\textbf{21.44}}      & \multicolumn{2}{c}{21.73}      \\ & \multicolumn{2}{c|}{SVHN}                                                                                                                                             & \multicolumn{2}{c|}{2.33}       & \multicolumn{2}{c}{\textbf{2.30}}       \\ \midrule \multirow{3}{*}{S3}                                                           & \multicolumn{2}{c|}{C10}                                                                                                                                              & \multicolumn{2}{c|}{2.50}       & \multicolumn{2}{c}{\textbf{2.49}}       \\ & \multicolumn{2}{c|}{C100}                                                                                                                                             & \multicolumn{2}{c|}{\textbf{21.05}}      & \multicolumn{2}{c}{21.08}      \\ & \multicolumn{2}{c|}{SVHN}                                                                                                                                             & \multicolumn{2}{c|}{2.32}       & \multicolumn{2}{c}{2.32}       \\ \midrule \multirow{3}{*}{S4}                                                           & \multicolumn{2}{c|}{C10}                                                                                                                                              & \multicolumn{2}{c|}{2.95}       & \multicolumn{2}{c}{\textbf{2.61}}       \\ & \multicolumn{2}{c|}{C100}                                                                                                                                             & \multicolumn{2}{c|}{21.48}      & \multicolumn{2}{c}{\textbf{20.66}}      \\ & \multicolumn{2}{c|}{SVHN}                                                                                                                                             & \multicolumn{2}{c|}{\textbf{2.44}}       & \multicolumn{2}{c}{2.45}\\ \bottomrule \end{tabular} } \end{center}
\vskip -0.3in
\end{wraptable}

\textbf{Robustness Validations}: To validate the robusteness of our adaptive scheme EDD, we elaborate another scheme i.e. \textbf{p}eriodic \textbf{c}yclic \textbf{d}ecay (PCD) which excluding EDD and determine $L^{\exp}$ statically before training as the additional tailored baseline. We first \textit{finetune the hyperparameters of both EDD and PCD on NB201\&C10 and then transfer the settings and recipes exactly to all other search spaces and datasets}. we finetune PCD even marginally surpass EDD in Table~\ref{table9}. On the contrary, by transferring the configuration of NB201\&C10, EDD starkly outperforms PCD on most other cases especially on NB201\&C100, DS\&C10 and DS\&IN-1k. This results underpin the robustness virtue of EDD over less flexible decay scheme PCD. Further investigation identify that PCD is more brittle than EDD to the warmup epoch and preseted decay strength on NB201 and DS respectively. We also note that simply increase these two values can recover the performance.

\begin{figure}[t]
\includegraphics[width=\columnwidth]{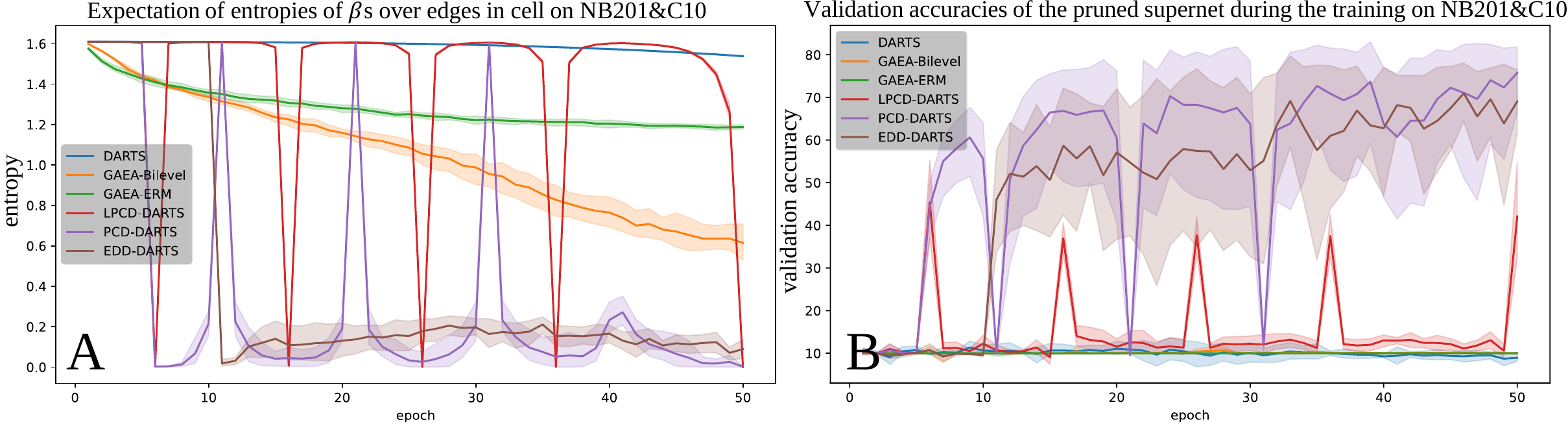}
\caption{(A). Trajectories of the expectation of entropies of $\beta$s during training on NB201\&C10. (B). Discretized accuracies on validation set in training on NB201\&C10.}
\label{fig4}
\vskip -0.2in
\end{figure}

\textbf{Analyses}:
According to \cite{li2021geometry}, the entropy of $\beta$ and the discretized accuracy of the pruned finalnet on validation set are two main measurements for evaluating the impact of the sparsity on method. We illustrate the dynamics of the expectation of entropies of $\beta$s over edges in space during the training on NB201\&C10 in Figure~\ref{fig4}A. in which the distribution entropy of EDD are much lower than other baselines. This validates our proposal of employing small temperature to sparsify $\beta$ in training. To better characterize the effect of the sparse $\beta$ on alleviating the mismatch, we also illustrate the corresponding discretized validation accuracies during the training on NB201\&C10 in Figure~\ref{fig4}B. 
Both DARTS and GEAE in Figure~\ref{fig4}B are basically stuck at the random discretized accuracies which is obviously due to the insufficient sparsity of parameters shown in Figure~\ref{fig4}A. 
In contrast, EDD-DARTS, shown by the brown line in Figure~\ref{fig4}B, maintains an appropriate sparsity of $\beta$ and steadily improve the discretized accuracies throughout the training. We observe the similar phenomenon on DARTS space. 
The drift of $E(a)$ can be seen in Figure\ref{fig4}A in which the three decay cycles of PCD exhibit slightly different  dynamics of $\beta$ for the same temperature sequence ${L^{\exp }}$. This is the downside of PCD since it determines the whole ${L^{\exp }}$ beforehand and fixes it during training. In contrast, EDD adaptively gets ${t^{\exp }}$ directly based on the sparsity depicted in Eq.(15), thereby finds the appropriate $t$ timely to compensate that drift of the expectation.

\textbf{Acknowledgment}: National Key R\&D Program of China (2022YFF0503900).

\textbf{Conclusion}:
In this paper, we focus on sparsifying the $\beta$ via utilizing and scheduling small temperature in DARTS. We first propose sn-softmax to alleviate the gradient saturation of the premature convergence. Next, we propose ETS to better control the sparsity of $\beta$ and we elaborate an entropy-based adaptive scheme EDD to finally deliver the effective enhancement in DARTS.

%
%
%
\bibliographystyle{splncs04}
\bibliography{mybibliography}

\end{document}